\documentclass[letterpaper, 10 pt, conference]{ieeeconf}  

\IEEEoverridecommandlockouts                              

\usepackage{hyperref}
\hypersetup{
	colorlinks   = true,
	citecolor    = gray
}

\title{\LARGE \bf Apple Counting using Convolutional Neural Networks
}

\author{Nicolai H{\"a}ni$^{1}$, Pravakar Roy$^{1}$ and Volkan Isler$^{1}$
\thanks{*This work was supported by the USDA NIFA MIN-98-G02}
\thanks{$^{1}$N. H{\"a}ni, P. Roy and V. Isler are with the Department of Computer Science \& Engineering,
        University of Minnesota, Minneapolis, MN 55455, United States of America
    {\tt\small \{haeni001, royxx268, isler\}@umn.edu}}%
}

\usepackage{amsmath}
\usepackage{algorithm2e}
\usepackage{graphicx}
\usepackage{subfig}
\usepackage{color}
\usepackage[dvipsnames]{xcolor}
\usepackage{dblfloatfix}

\usepackage{cite}

\graphicspath{ {figures/} }

\begin{document}
\maketitle
\thispagestyle{empty}
\pagestyle{empty}

\begin{abstract}
Estimating accurate and reliable fruit and vegetable counts from images in real-world settings, such as orchards, is a challenging problem that has received significant recent attention. Estimating fruit counts before harvest provides useful information for logistics planning. While considerable progress has been made toward fruit detection, estimating the actual counts remains challenging. In practice, fruits are often clustered together. Therefore, methods that only detect fruits fail to offer general solutions to estimate accurate fruit counts. Furthermore, in horticultural studies, rather than a single yield estimate, finer information such as the distribution of the number of apples per cluster is desirable.
In this work, we formulate fruit counting from images as a multi-class classification problem and solve it by training a Convolutional Neural Network. We first evaluate the per-image accuracy of our method and compare it with a state of the art method based on Gaussian Mixture Models over four test datasets. Even though the parameters of the Gaussian Mixture Model based method are specifically tuned for each dataset, our network outperforms it in three out of four datasets with a maximum of 94\% accuracy.  Next, we use the method to estimate the yield for two datasets for which we have ground truth. Our method achieved 96-97\% accuracies. For additional details please see our video here: \href{https://www.youtube.com/watch?v=Le0mb5P-SYc}{https://www.youtube.com/watch?v=Le0mb5P-SYc}.
\end{abstract}
\section{Introduction}
Precision agriculture methods are starting to change how food is produced in our world. 
There are now commercial solutions for numerous farming tasks such as monitoring crop fields, automatic seeding, harvesting, and packaging. However, the application domain of precision agriculture has been primarily limited to commodity crops such as wheat, rice or maize. Developing solutions for specialty crops (such as fruits or vegetables) has been challenging due to the complex geometry of orchards compared to row crops. Nevertheless, significant research in recent years has been dedicated to automating farming tasks for specialty crops. In this work, we address a specific precision agriculture task: counting apples from images.

Estimating accurate fruit counts at the image level enables precise yield estimation and mapping. Availability of these metrics enables farmers to monitor the current state of their orchard, plan for optimal utilization of their workforce and make informed decisions before the harvesting process. In the absence of an automated system, these counts are currently estimated by manually sampling a few trees at random and extrapolating these counts to the whole orchard, which is inaccurate. Counting information is also valuable for horticultural studies where scientists collect phenotype data associated with yield. 

Although there has been significant progress towards yield estimation by detecting single fruits under regular environmental conditions, the problem itself remains challenging for the following reasons: (1)~Fruits and vegetables come in all shapes, sizes, and colors. Even when only concentrating on a particular fruit, such as apples in our case, there is still a significant color variety in the data. (2)~Fruits often grow in clusters of arbitrary size and shape. (3)~They are often occluded either by other fruits, branches or leaves.

\begin{figure}[h]
	\centering
	\subfloat[][ Given an image and detected fruit]{{\includegraphics[width=0.45\columnwidth]{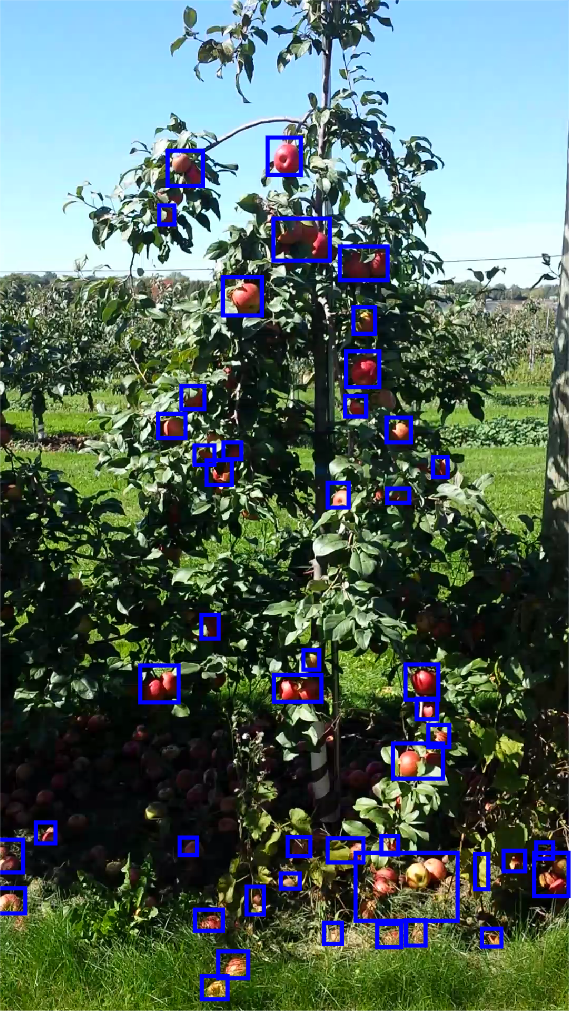}\label{fig:intro1}}}%
	\quad \quad
	\subfloat[][Determine fruit counts ]{{\includegraphics[width=0.45\columnwidth]{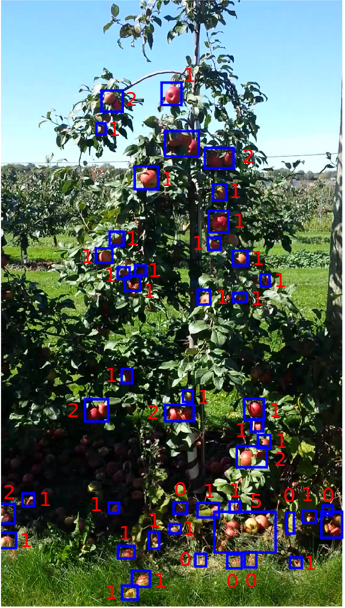}\label{fig:intro2}}}%
	
	\caption{Input and output of the proposed fruit counting algorithm}
	\label{fig:intromain}
\end{figure}

In this paper, the main question we would like to answer is: Given an image and detected clusters of fruits, can existing neural network architectures be leveraged to count the fruits? An example is shown in Fig.~\ref{fig:intromain}.

\begin{figure*}[!t]
	\centering
	\includegraphics[width=\textwidth]{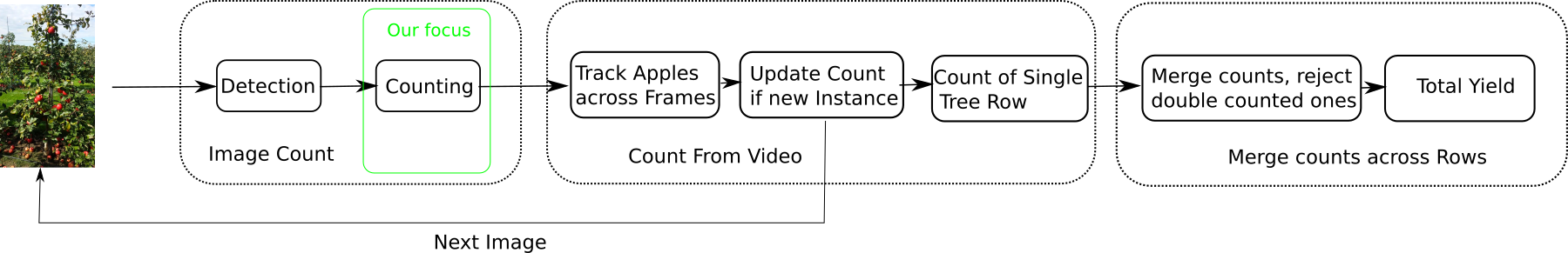}
	\caption{Steps for automated yield estimation}
	\label{fig:main1}
\end{figure*}

In this paper, we show that counting of clustered fruit can be formulated as a classification problem. Our method takes images and detected regions of fruits clusters as inputs and predicts a single count per cluster. Once the model is learned we are able to predict correct counts in up to $94\%$ of the cases. The network is evaluated on data of large variety: (1) Fruit colors vary from red to green; (2) Data contains fruits on trees and on the ground; (3) Data were collected from different tree rows and throughout the year; (4) Illumination conditions vary from cloudy to sunny; (5) data were collected at varying distances between the sensor and trees. 
The key contributions are: 
\begin{itemize}
	\item A method to predict fruit counts from a single look at an image patch without the need of prior segmentation, fine-tuning of any parameters or a subsequent algorithm for counting. 
	\item An analysis of image-based counting performance over four datasets as well as an evaluation of our method on the task of complete yield estimation. We provide one of the first studies that use multiple, RGB-only datasets from \emph{multiple tree varieties and multiple years} to test the performance of deep learning methods on the counting task.
\end{itemize}

The rest of this paper is organized as follows: Section \ref{sec:relwork} presents related work on the topic of yield estimation. The problem is defined in Section \ref{sec:main}. Experiments together with strengths and weaknesses are presented in Section \ref{sec:results}. In Section~\ref{sec:discussion} we discuss our results and future work.
\section{Related Work}
\label{sec:relwork}

Various approaches have been proposed to count fruits, both in indoor and outdoor environments. The proposed methods differ in the sensors used (RGB-\cite{changyi_apple_2015, chen_counting_2017, kulic_vision-based_2017, bargoti_deep_2017}, thermal-\cite{wachs_low_2010}, and spectral- cameras~\cite{okamoto_green_2009} as well as LIDAR~\cite{stein_image_2016}) and  in the amount of control that is exercised over the test setup. Multiple systems have tried to restrict measurements to specific times of the day~\cite{wang_automated_2013,linker_machine_2018}. Some rely on using a mobile canopy that is put over the crop rows~\cite{gongal_apple_2016}. Such methods are costly and difficult to apply. Ideally, a solution to the fruit counting problem should be platform independent and not rely on special sensors nor controlled environments.

\subsection{Fruit Detection}
Counting fruits by detection is a well-studied problem in the literature. Early systems were dominated by hand-crafted color thresholds~\cite{wang_automated_2013, ji_automatic_2012}. More recently, researchers have leveraged the predictive power of modern machine learning techniques to learn a combination of features to robustly detect fruits. Advances in deep learning have led to the adoption of neural networks for fruit detection. Zhou et al.~\cite{changyi_apple_2015} used a back propagation network to classify the image into apple and background pixels. After segmentation, they used edge detection and Hough Transform to detect single apple instances that are summed up in order to obtain a count. Bargoti and Underwood~\cite{bargoti_image_2017} used Multi-Scale Multi-Layered Perceptrons (MLP) together with a CNN for apple and almond detection. Stein et al.~\cite{stein_image_2016} proposed a multi-sensor framework, consisting of a camera and LIDAR for mango counting. By combining these two sensors and through using a multi-view approach for fruit tracking and counting, they achieved counting accuracies of $\sim 98\%$. It is important to note, that mangoes do not grow in clusters like apples, which makes counting by detection a feasible approach for mangoes.

More recently, Faster R-CNN~\cite{ren_faster_2017} integrated region proposal and classification step into a single network. This approach has already been adapted to fruit and vegetable detection. Bargoti et al.~\cite{bargoti_deep_2017} used such a network to detect and count almonds, mangoes and apples, achieving F1-scores of $>0.9$. Sa et al. \cite{sa_deepfruits:_2016} used a similar approach to detect green peppers. They merged RGB and Near Infrared (NIR) data and achieved F1-scores of $0.84$. In this paper, we use image data from a single RGB camera in cases where modern detection networks fail to count fruits accurately. Specifically, we want to accurately count clustered apples. 

\subsection{Counting of Clustered Fruit}
While the detection part of the processing pipeline has seen rapid advances that culminated in the adoption of Faster R-CNN in orchards, the counting part has largely been neglected. Current methods to receive counts from segmented images are dominated by either Circular Hough Transform (CHT)~\cite{gongal_apple_2016, silwal_apple_2014, changyi_apple_2015} or Non-Maximum Suppression(NMS)~\cite{bargoti_deep_2017, sa_deepfruits:_2016}. The main drawbacks of CHT are its reliance on accurate segmentation of the image and the need to fine tune parameters across datasets. NMS relies on a static threshold to reject overlapping instances and counts are generated by summing up the filtered bounding boxes.

Chen et al.~\cite{chen_counting_2017} used a combination of two networks for counting. The first, fully convolutional network creates activations of possible targets in the image, while the second network counts these targets. Rahnemoonfar and Sheppard~\cite{maryam_rahnemoonfar_deep_2017} use a CNN for direct counting of red fruits from simulated data. They used synthetic data to train a network on 728 classes to count red tomatoes. In previous work, Roy and Isler~\cite{kulic_vision-based_2017} used Gaussian Mixture Models (GMMs) and Expectation Maximization (EM) to accurately count apples from clusters. This method achieves state-of-the-art counting performance of up to $91\%$, but it relies on accurate image segmentation and is prone to errors due to wrongly segmented pixels. In this paper, we show that modern deep learning based techniques can achieve accurate cluster counts without the need of prior segmentation or additional post-processing.
\section{Method}
\label{sec:main}

\subsection{Motivation}
\label{sec:problem}
A fully autonomous system for yield estimation in orchards usually follows the illustration in Fig.~\ref{fig:main1}.
The system receives images as inputs. Apple clusters are detected in the image and a counting algorithm computes per image counts. In general, apple clusters will be visible in more than just a single image. To avoid double counting, the individual clusters have to be tracked and counts have to be monitored. At the end of a sequence, we have one count per tree row. For complete yield estimates the front and back side of a corresponding tree row has to be merged. Again double counting has to be avoided.

\begin{figure}[b]
	\centering
	\subfloat[][Output of Faster R-CNN]{{\includegraphics[width=0.45\columnwidth]{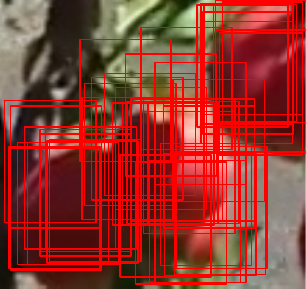}\label{fig:main2a}}}%
	\quad \quad
	\subfloat[][Output after Thresholding and Non-maximum suppression]{{\includegraphics[width=0.45\columnwidth]{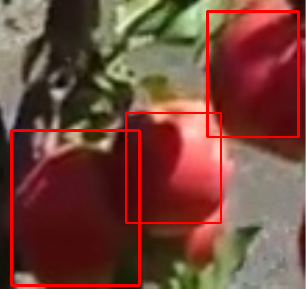}\label{fig:main2b}}}%
	\caption{Case of under counting of Faster R-CNN}
	\label{fig:failureRCNN}
\end{figure}

The detection part has received significant attention recently and with Faster R-CNN we can detect single fruits accurately~(\cite{bargoti_deep_2017}, \cite{sa_deepfruits:_2016}). However, these approaches fail when fruits are growing in clusters, where one fruit is partially occluded by others, as seen in Fig.~\ref{fig:failureRCNN}. In this case, the network generates 2000 region proposals for image patches of $500\times 500$ pixels, which are moved over the image with stride 50 pixels~\ref{fig:main2a}. All the proposals are then merged together and filtered using Non-maximum suppression with an Intersection over Union (IoU) threshold of 0.7~\ref{fig:main2b}. In this case, the algorithm will miss one of the apples in the cluster.  Bargoti and Underwood~\cite{bargoti_deep_2017} found that their approach contains errors of $\sim4\%$ due to wrongly counting clusters.
Roy and Isler~\cite{kulic_vision-based_2017} used a GMM based approach to count apples in clusters. Their semi-supervised detection method was $96-97 \%$ accurate. For computing the counts from all detected clusters they achieved accuracies of $80-91\%$. Most of the discrepancy between detection and counting accuracies were attributed to either over- or under-counting due to wrong segmentations.

\subsection{Proposed Approach}
We approach the problem of accurately estimating apple counts by making the following observations: (1) Apples are sparsely distributed over the whole image; (2) they are often clustered together; (3) and cluster sizes are unevenly distributed. From these observations, we define the accurate counting of clustered apples as a classification problem with a finite number of classes representing counts of apples per image patch. In our approach, we do not restrict the total count per image, only the maximum cluster size. We also do not restrict the model to only red fruits. Instead, we demonstrate that our approach generalizes across varying orchards and fruit colors. 

\subsection{Detection of Apple Clusters}
\label{sec:detection}
Our method starts with the output of a region proposal algorithm which can be any method that can generate candidate locations for apple clusters. We use a color-based segmentation method reported in our previous work~\cite{kulic_vision-based_2017}. We briefly describe it here for completeness.


The detection algorithm takes a color image and generates bounding boxes of probable apple clusters. In the first step, the image is over-segmented into superpixels in the LAB colorspace. Each of these superpixels is then represented by its mean color. These superpixels are clustered into $25$ color classes. We use a user supervised training method to collect relevant color clusters that represent apple clusters over multiple data sets. Finally, we use Kullback-Leibler (KL) divergence to classify each superpixel into apple or background. We generate a binary mask from which we get the bounding boxes of probable apple clusters. Since this method uses color as the sole feature to classify pixels into the corresponding clusters, it is prone to detect false positives. Therefore, it is crucial that the counting algorithm rejects false detections.

\subsection{Counting with Deep Neural Networks}
The input to the network is the original color image and detected apple clusters. It predicts a count for each of the probable apple cluster regions, detected by the region proposal algorithm. The size of these apple clusters is not arbitrarily large, usually between one to six apples. We, therefore, define 7 classes, representing the apple counts per image patch (including zero). 

In this work, we used the older, shallower network architecture of AlexNet for a proof of concept. Our proposed network consists of the AlexNet convolutional layers initialized with the ImageNet weights. We removed all the fully connected layers and add back three new ones with the dimensions of our problem. We used dropout layers to counter overfitting. We kept the original image input layer of AlexNet which requires us to upsample images to $227\times227$ pixels. We used transfer learning followed by a  fine-tuning step: First we only trained the newly added layers for 5 epochs. Then we trained all the convolutional layers as well, but we adopted the per layer learning rate to $0.0001$. We trained the full network for up to 30 epochs. 

\subsection{Implementation Details}
We trained our model on a desktop computer with a single NVIDIA GTX 1080 GPU with 8GB of memory. The network was implemented in Matlab. For the optimization of the network, stochastic gradient descent was used. The initial learning rate was set to $0.001$, momentum to $0.9$ and the learning rate decay was $0.2$ which we applied every 5 epochs during the fine tuning step. For image augmentations we followed Taylor and Nitschke~\cite{taylor_improving_2017} and \cite{bargoti_deep_2017} who studied the effects of a variety of image augmentation methods. They found that the most effective ones are random cropping, flipping and color jittering. We did not use random cropping, as these transformations change the true label associated with the image.
\section{Experiments}
\label{sec:results}
In this section, we present the experimental setup including datasets, and report image patch counting results and share some insights.

\subsection{Datasets}
We collected multiple datasets to validate our approach. For this paper, the data was split into training-, validation-, and yield estimation categories. For training, we used images collected from multiple orchards in Minnesota and Washington. All of our evaluation data was collected at the University of Minnesota Horticultural Research Center (HRC) in Eden Prairie, Minnesota between the years 2015 and 2016. Since this is a university orchard used for phenotyping research, it is home to many different kinds of apple tree species. We collected video footage from different sections of the orchard using a standard Samsung Galaxy S4 cell phone. During data collection, video footage was acquired from which the images were extracted. The reason we use the HRC data is that we were also able to collect ground truth information post-harvest.

\subsubsection{Training Datasets}
For training, we used two datasets, one containing green and one containing red apples. Both datasets were obtained from the sunny side of the tree row. From these two datasets, we extracted image patches using the method described in Section~\ref{sec:detection}. In total, we got $13000$ image patches which were annotated manually. Additionally, we extracted $4500$ patches at random that do not contain apples. To balance our training set between classes we up-sampled the training dataset, using small rotations of $\pm 5\deg$ and data augmentation, to a total of $64000$ image patches.

\subsubsection{Single Frame Validation Datasets}
\begin{figure}[t]
	\centering
	\subfloat[]{{\includegraphics[width=0.24\columnwidth]{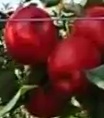}}}%
	\quad
	\subfloat[]{{\includegraphics[width=0.21\columnwidth]{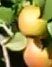}}}%
	\quad
	\subfloat[]{{\includegraphics[width=0.21\columnwidth]{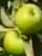}}}%
	\quad
	\subfloat[]{{\includegraphics[width=0.18\columnwidth]{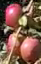}}}%
	\caption{Sample images from datasets 1-4. \label{fig:res1}}
\end{figure}

To validate our image patch based counting approach we evaluated on four different datasets. Similarly to the training dataset, we used the method described in Section~\ref{sec:detection} to extract image patches, that were then annotated manually. Fig.~\ref{fig:res1} shows example patches of the four datasets. The details of these four datasets are as follows:\\

\textbf{Test Set (a)} This dataset contains 5054 patches of red apples. The data set is a collection of red patches from both sides (sunny/shady) of multiple rows. The sunny side of the row shows false positive detections since the dataset was acquired in September and the leaves started to turn yellow.\\
\textbf{Test Set (b)} This dataset contains 599 patches of mainly yellow and orange apples in the early growth phase. It was acquired on a sunny day and most of the images show apples in bright sunlight.\\ 
\textbf{Test Set (c)} This dataset contains 574 images of a green apple variety. The dataset contains apples in the shadow as well as in bright illumination.\\
\textbf{Test Set (d)} This dataset contains 704 patches showing red apples again, but this time the data was acquired from a larger distance. The apples appear smaller and at a lower resolution. Apples appear both in bright sunlight and in the shadow.

\subsubsection{Yield Estimation Datasets}
\begin{figure}[b]
	\centering
	\subfloat[Yield validation set 1 example image]{{\includegraphics[width=0.45\columnwidth]{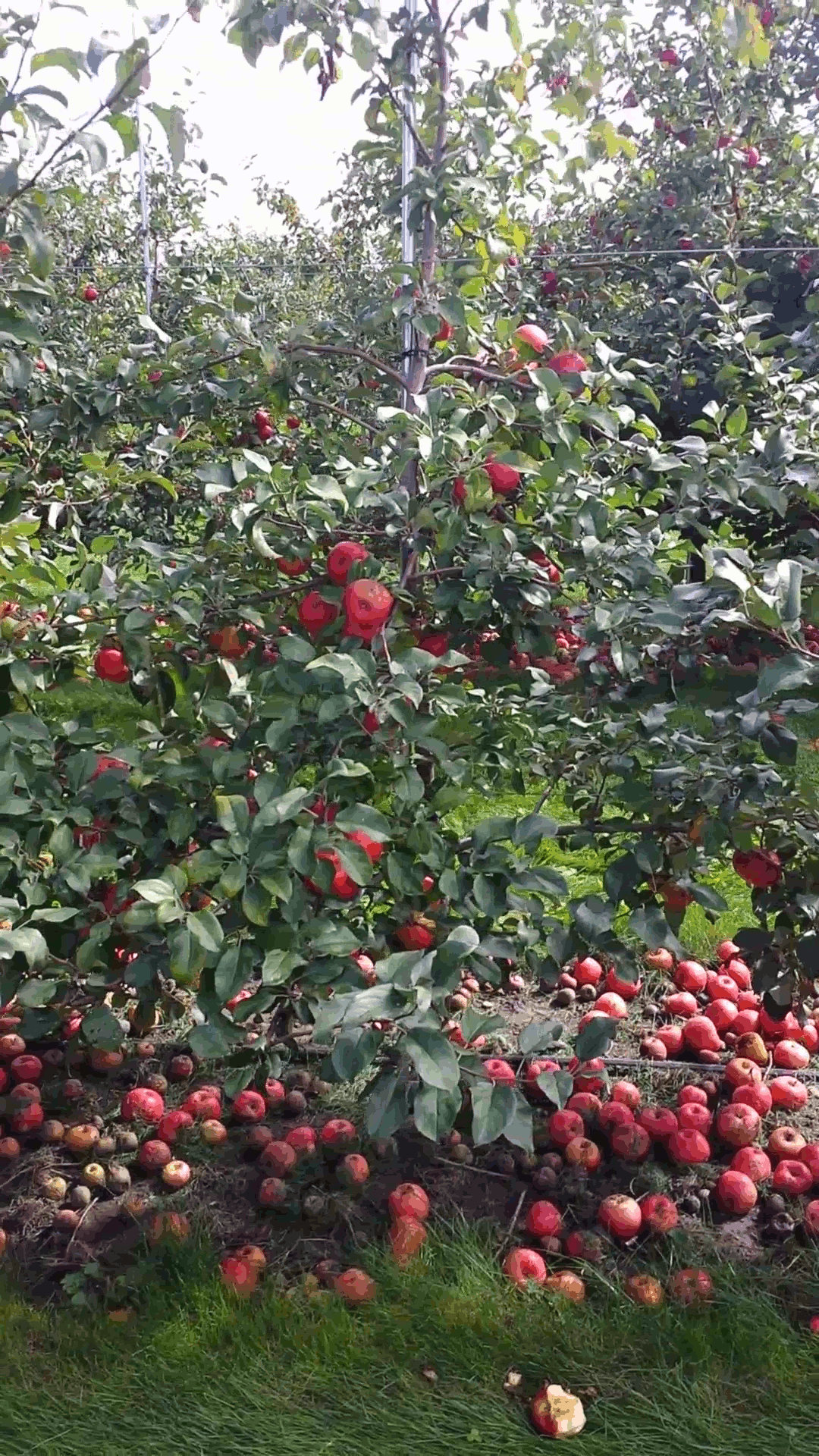}\label{res:d1} }}%
	\quad \quad
	\subfloat[Yield validation set 2 example image]{{\includegraphics[width=0.45\columnwidth]{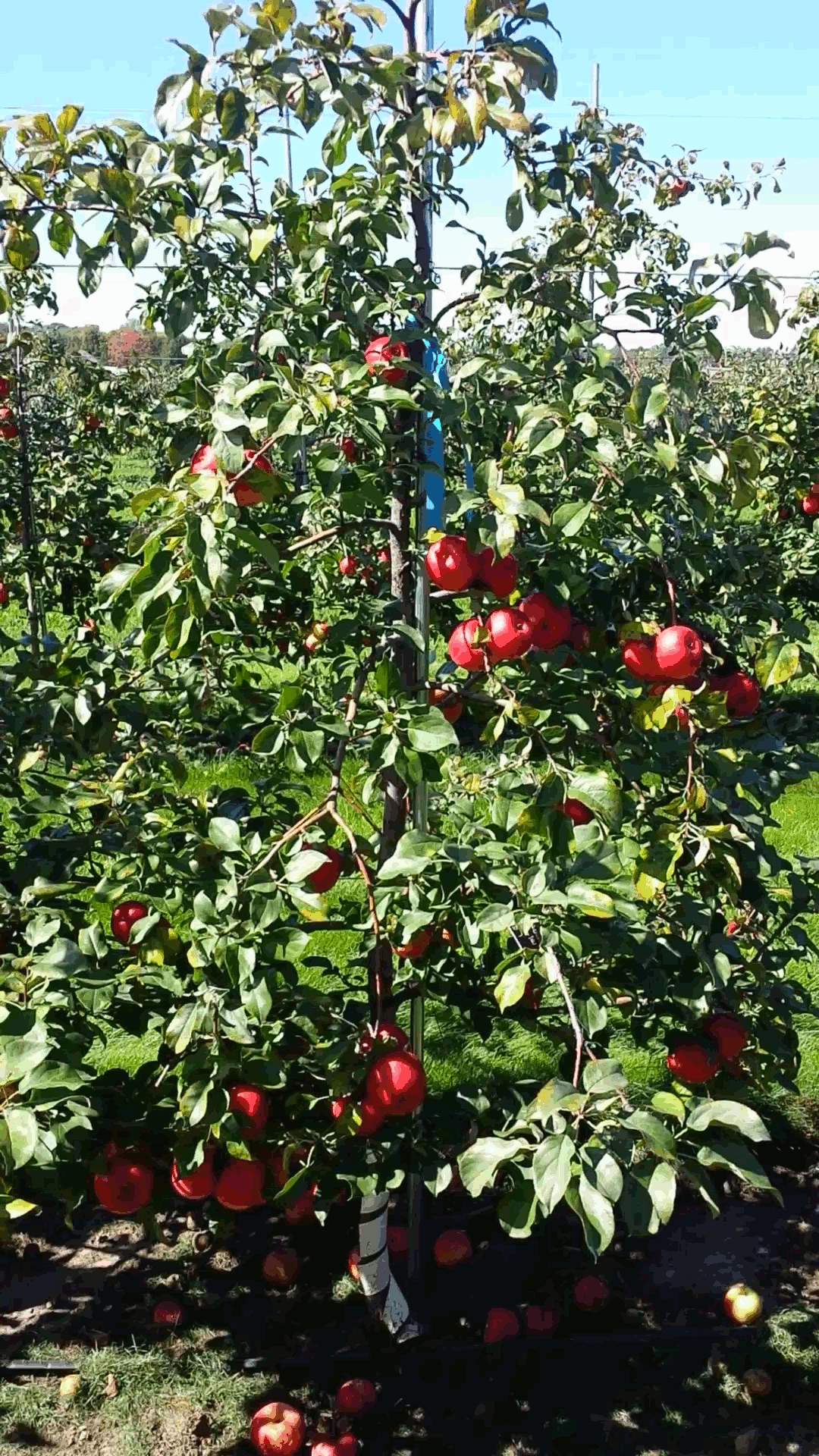}\label{res:d2}}}%
	\caption{Used yield estimation datasets}
\end{figure}

We use two datasets to show how our approach performs on estimating yield.\\
\textbf{Yield validation set 1:} This dataset contains six trees in a row with a total of 270 apples. Fig.~\ref{res:d1} shows an example image.\\
\textbf{Yield validation set 2:} This data set contains eight trees in a row with a total of 274 apples. Fig.~\ref{res:d2} shows an example image.\\

We obtained two videos for each of these datasets, one from the front and one from the backside of the tree row. Ground truth counts were obtained from the harvest.

\subsection{Image Patch Counting Results}
Table~\ref{tab:1} shows the results of the cluster counting performance. We present counting results in comparison with the GMM approach presented in~\cite{kulic_vision-based_2017} since we have this method available and it achieves state-of-the-art detection and counting performances. 

The neural network outperforms the GMM model on all but one of the test sets. On test set 3 we outperform the GMM by almost $15\%$. The CNN is able to generalize between datasets even under varying illumination conditions. On the test sets 2 and 4 we outperform the GMM by $12\%$ and $10\%$ respectively. However, for test set 1, the GMM performs better by $9.6\%$. In retrospect, we realized that this dataset contains many apples on the ground which appear in configurations very different from what's available in the training dataset. We attribute the low performance of the proposed approach to this discrepancy.

\begin{table}[t!]
	\begin{center}
		\caption{Image Patch Counting Results}
		\label{tab:1}
		\begin{tabular}{|c|c|c|c|c|}
			\hline
			\textbf{Approach} & \textbf{Test Set 1} & \textbf{Test Set 2} & \textbf{Test Set 3} & \textbf{Test Set 4}\\
			\hline
			GMM & \textbf{89.8 \%} & 81.8 \% & 79.9 \% & 76.7 \% \\
			\hline
			CNN & 80.21 \% & \textbf{93.32 \%} & \textbf{94.3  \%} & \textbf{86.1 \%}\\
			\hline
		\end{tabular}
	\end{center}
\end{table}

\begin{figure}[b]
	\centering
	\includegraphics[width=\columnwidth]{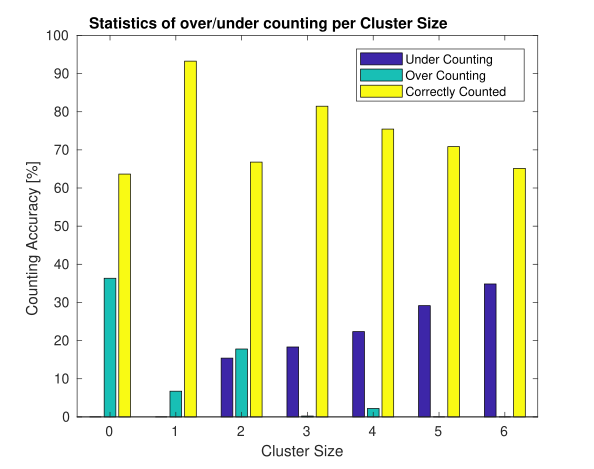}
	\caption{Counting performance across different cluster sizes}
	\label{fig:4}
\end{figure}

Fig. \ref{fig:4} shows the performance of our network over different cluster sizes. We realize that for false positive detections (zero counts) our method overcounts compared to ground truth in $36 \%$ of the instances. With increasing cluster size we start to undercount the actual number of fruits more heavily (up to $34\%$ for clusters of six apples). However, the network does rarely over-/ or undercount by a single apple.

\subsection{Yield Estimation Results}
The neural network gives us a count per image frame, composed of the sum of all patch counts. However, this in itself does not directly correspond to yield. To provide yield data we need to integrate counts over multiple views throughout the dataset. To avoid double counting of fruits we need to track the individual regions across the entire sequence. In this section, we provide a validation method for our approach to the application of yield estimation. We note that we only use this method for empirical validation of the counting approach, not as a comprehensive solution to the yield mapping problem.

\begin{figure}[t]
	\centering
	\subfloat[Shady side of tree row]{{\includegraphics[width=0.52\columnwidth]{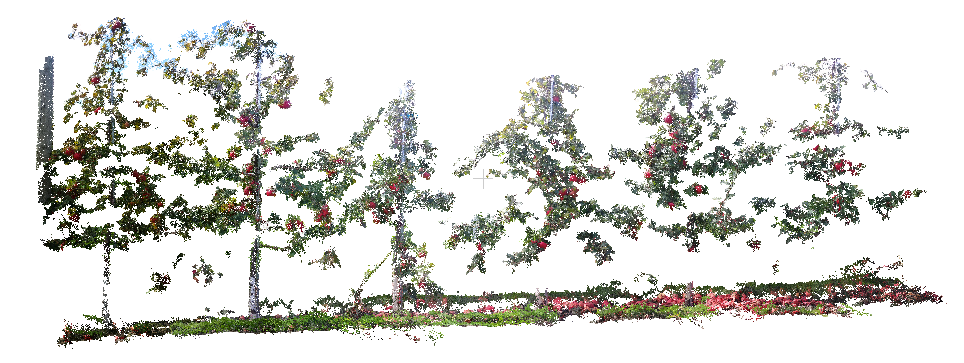}\label{fig:shady} }} \quad
	\subfloat[Sunny side of tree row]{{\includegraphics[width=0.4\columnwidth]{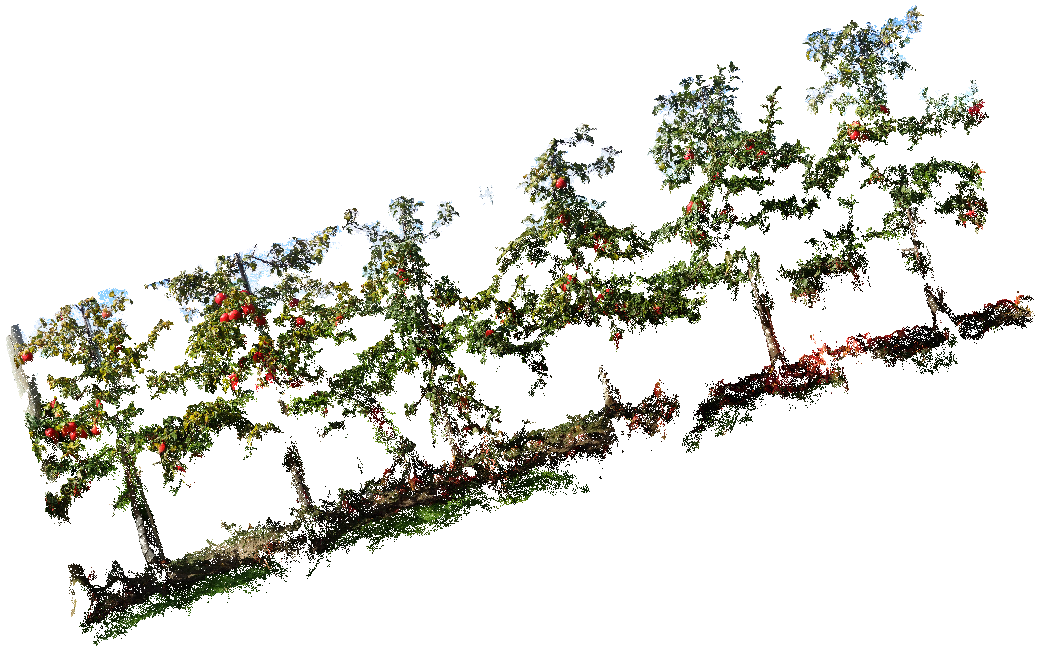}\label{fig:sunny} }} 
	\\
	\includegraphics[width=\columnwidth]{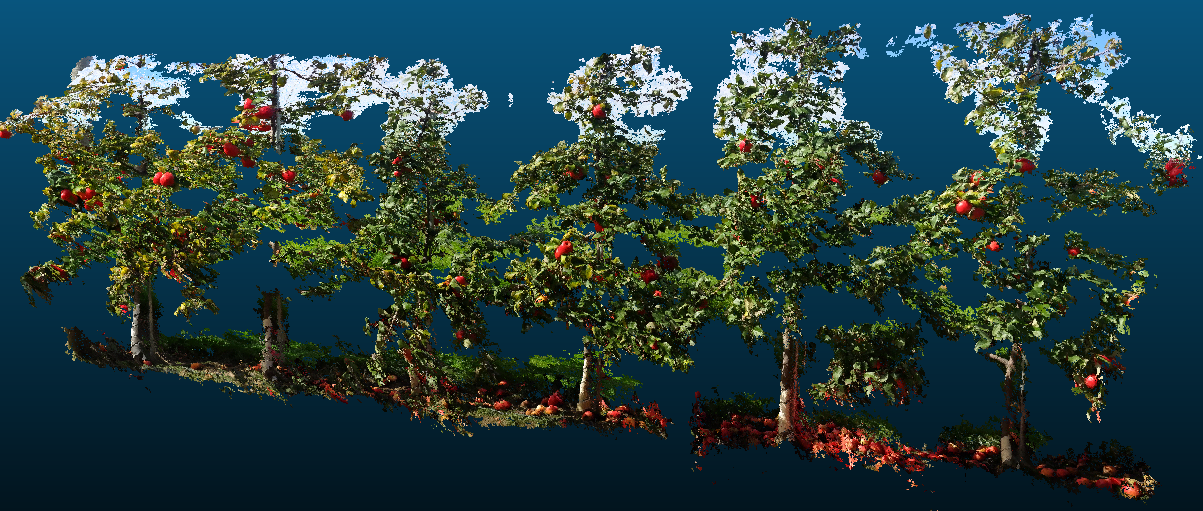}
	\caption{Merged 3D reconstruction of yield validation set 2}
	\label{fig:reconst}
\end{figure}

In cluttered environments, such as apple orchards, the appearance of any given cluster will change drastically between views. It is possible that in some frames, the apples are partially occluded or not visible at all. For this reason, we need to merge the counts across frames. In addition to tracking fruits between images, we need to avoid double counting when scanning a tree row from the front and from the back.
For our test setup, we first reconstruct both, the sunny and shady side of the tree row using the Agisoft software package as seen in Fig.~\ref{fig:shady}~\ref{fig:sunny}. We merge both reconstructions as seen in Fig.~\ref{fig:reconst}, using the algorithm of Roy et al.~\cite{roy_registering_2018}. 

\begin{figure}[t]
	\centering
	\includegraphics[width=\columnwidth]{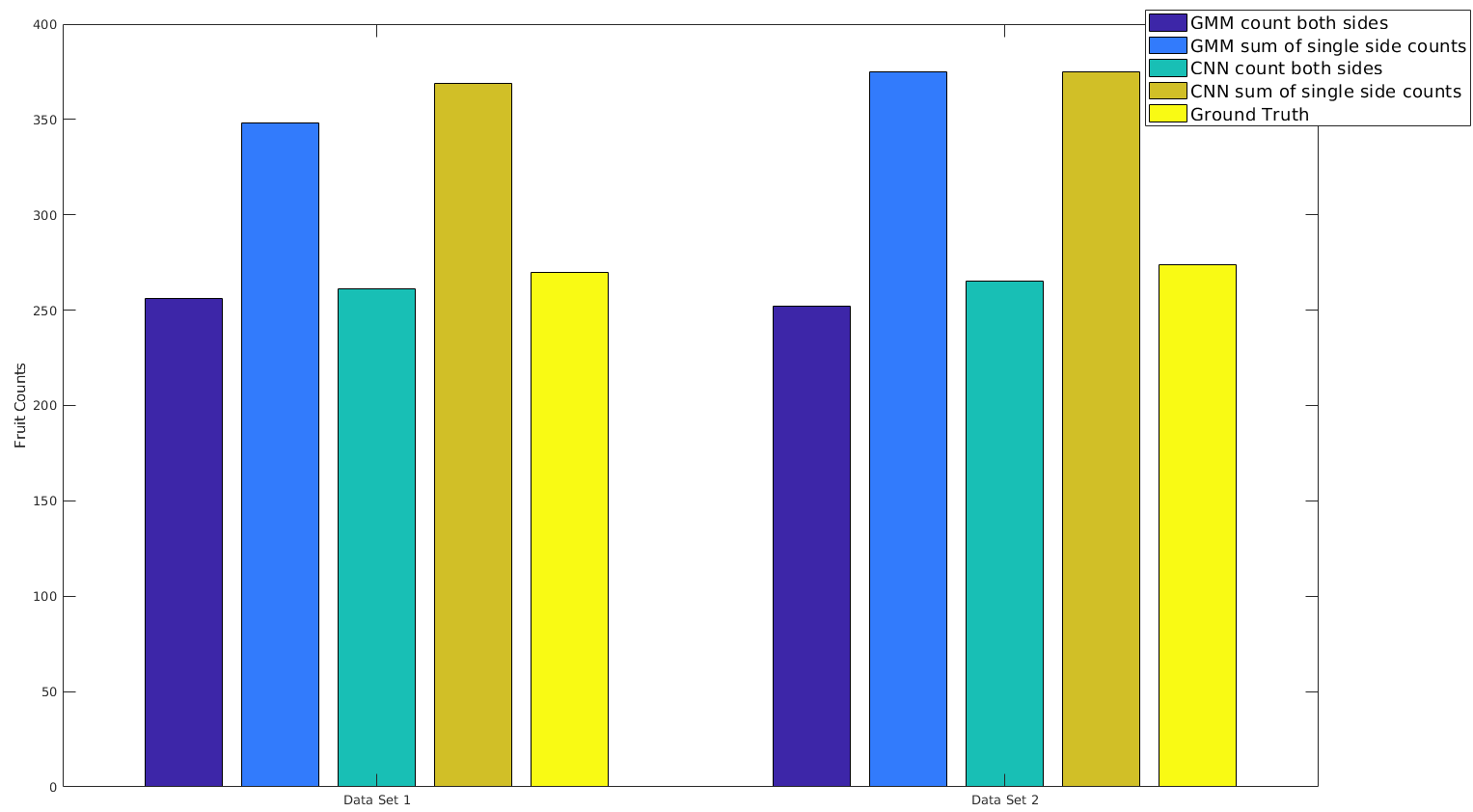}
	\caption{Yield estimation results}
	\label{fig:bothsides}
\end{figure}

We use the method of~\cite{kulic_vision-based_2017} to detect all single instances of fruits in both video sequences. The detections are back-projected into the 3D reconstruction to obtain fruit locations in 3D. A connected component analysis is performed in 3D to get individual clusters. Finally, image patches are generated by projecting individual clusters into the frames from which they are visible. 
The proposed counting network and the GMM based method are used to generate per patch counts on the projected regions. We take the three maximum predictions of corresponding patches and report the mean count.
In a post-processing step, all apples lying on the ground are removed using a height threshold. To merge the counts from both sides, they were summed up over all connected components. The intersection areas among the components are computed and we add/subtract the weighted parts accordingly. Fig~\ref{fig:bothsides} shows the results.

The accuracy of the proposed method is $96.97\%$ in dataset 1 and $96.72\%$ in dataset 2. The performance of the GMM based approach is $94.81\%$ and $91.97\%$ respectively.

\subsection{Qualitative Analysis of Failure Cases}
\begin{figure}[h!]
	\centering
	\subfloat[]{{\includegraphics[width=0.2\columnwidth]{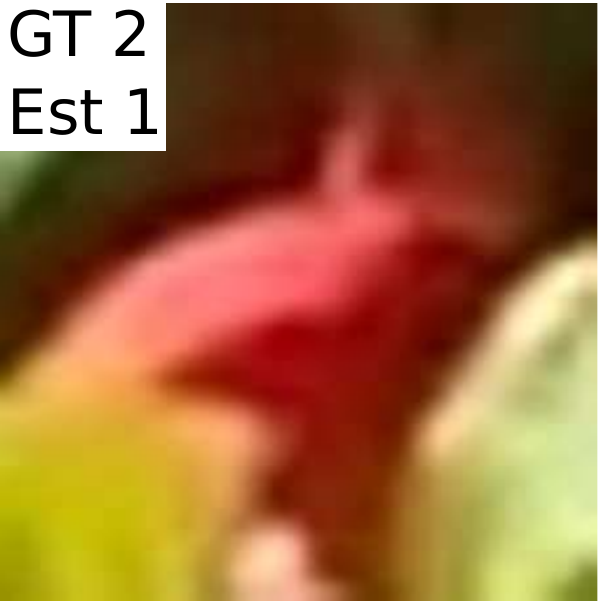}\label{5a} }}%
	\quad
	\subfloat[]{{\includegraphics[width=0.2\columnwidth]{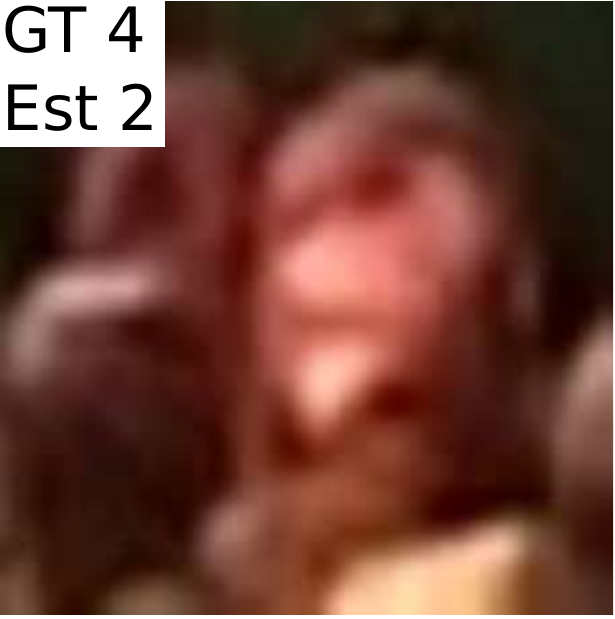}\label{5b} }}%
	\quad
	\subfloat[]{{\includegraphics[width=0.2\columnwidth]{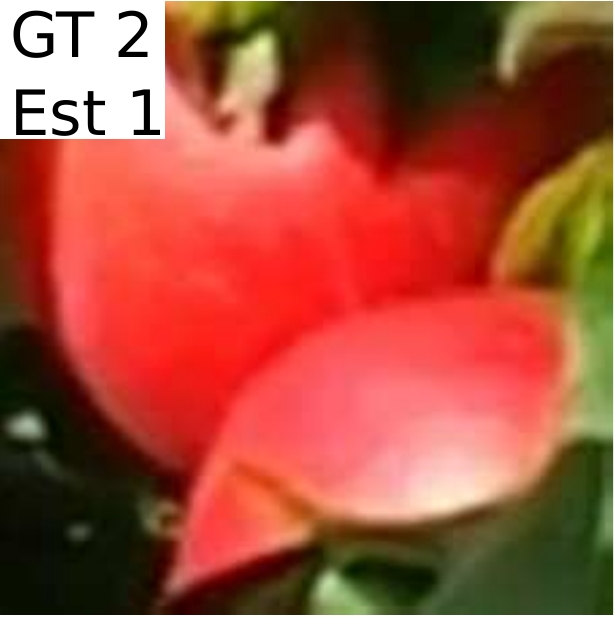}\label{5c} }}%
	\quad
	\subfloat[]{{\includegraphics[width=0.2\columnwidth]{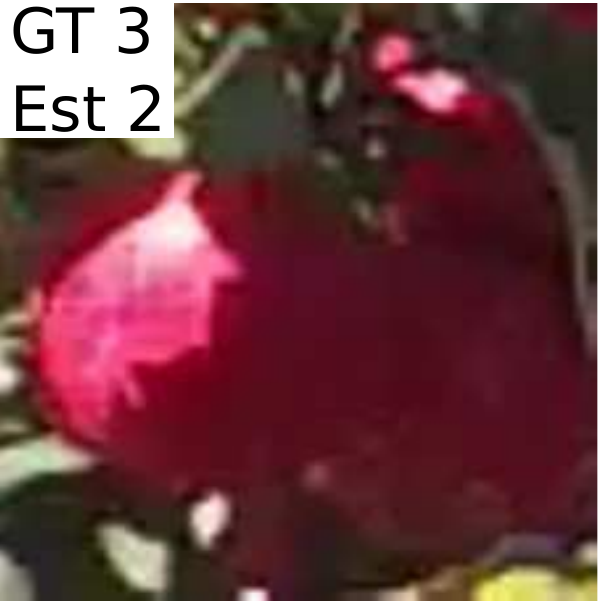}\label{5d} }}\\
	
	\subfloat[]{{\includegraphics[width=0.22\columnwidth]{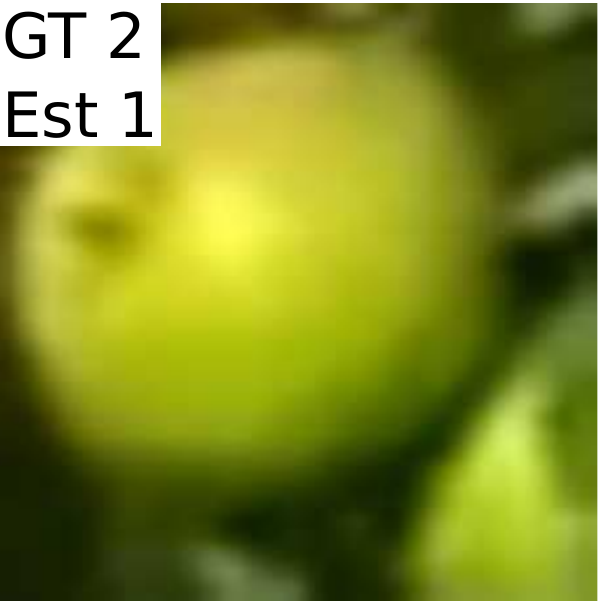}\label{5e} }}%
	\quad
	\subfloat[]{{\includegraphics[width=0.16\columnwidth]{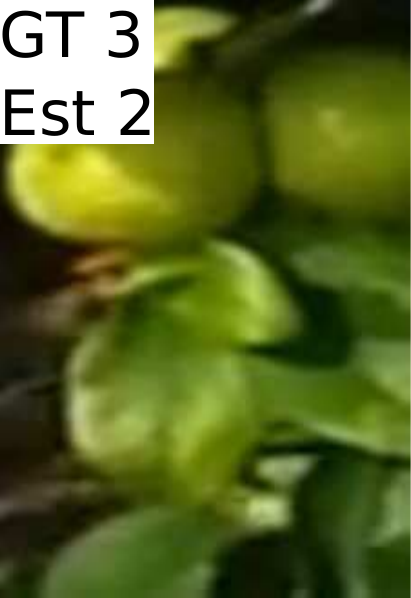}\label{5f} }}%
	\quad
	\subfloat[]{{\includegraphics[width=0.22\columnwidth]{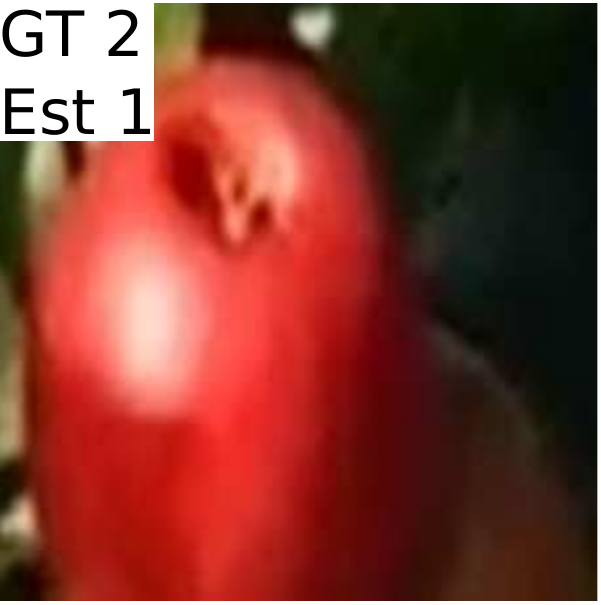}\label{5g} }}%
	\quad
	\subfloat[]{{\includegraphics[width=0.22\columnwidth]{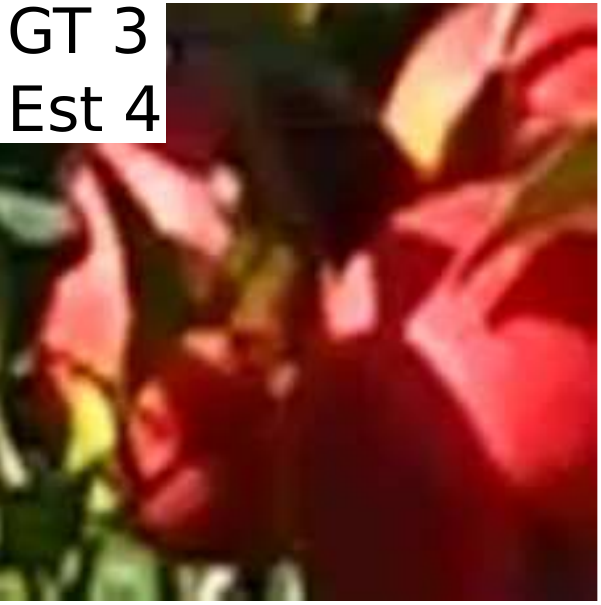}\label{5h} }}\\
	\caption{Some example failure cases of our method}%
	\label{fig:5}%
\end{figure}

In Fig.~\ref{fig:5} we show some failure cases of our method. In the original images, we noted the ground truth counts as \textit{GT} and the networks estimates as \textit{Est}.

The most common type of error is the one seen in Fig.~\ref{5a}, \ref{5c}, \ref{5e} and \ref{5g}. Here the images clearly contain multiple apples, but due to occlusions by other fruit or leaves, only a small portion of the fruits is visible in the image. In all of these cases, the second fruit is too small. This error is mainly due to incorrect selection of the regions during detection.
The error shown in Figure~\ref{5b} shows brown and yellow ground apples in the shady part of the original image. The network can only pick out the two most significant features, with the rest being treated as image noise. This failure case is prevalent in test set 1 but does not occur in the others. This type of error can be avoided by excluding all apples on the ground during the detection phase, or by adding examples of ground apples to the training dataset.
For the error case shown in Figure~\ref{5d} even human labelers cannot give an accurate count with $100\%$ certainty. The fruits form a dense cluster with little to distinguish between them. Our method finds two of the three fruits.
The error shown in Figure~\ref{5f} shows a ground truth label which has been annotated incorrectly. Even when taking special care during the labeling process these cases occurred often. However, by using synthetically rendered data this error could be eliminated. 
Lastly, we see an example of strong occlusion effects in Figure~\ref{5h}. The fruits are occluded by leaves which cuts one of them in two halves.

\section{Conclusion}
\label{sec:discussion}
In this paper, we addressed the problem of accurately counting clustered apples directly from images. 
We presented a method based on AlexNet, which we modified and fine-tuned on our own training data. We presented results which show that the method is more accurate than the previous, GMM based approach in three out of our four test data sets. Our multi-class classification approach achieves accuracies between $80\%$ and $94\%$ without the need of any pre- or post-processing steps. The deep learning network presented in this paper generalizes across different fruit colors,  occlusions, and varying illumination conditions.
The method was further evaluated on two rows to test its suitability for yield estimation. The fruits were counted from the front and back side of the tree row individually before merging them. Our approach achieved $96.7\%$ and $96.9\%$ accuracy with respect to ground truth obtained by counting harvested apples.

The CNN outperformed the GMM based method in three out of the four test sets. The network generalizes across varying illumination conditions, partial occlusions of fruits and fruit colors.
The result on the test set 3, which contains green apples compared to the red data sets is especially impressive. One reason might be that the variance in color among green apples in the training and test sets is much smaller than for red apples. The network is able to distinguish very clearly between leaves and apples, even if leaves turn brown in September (as was the case in test set 1).
After visually inspecting test set 1 we note that it contains many clusters of apples on the ground that are misclassified by the network (See Figure~\ref{fig:intromain}). Since the yield dataset contains similar colored apples as test set 1 we compared data and results of the two. In test set 1 we achieved $80\%$ accuracy, while on the yield data we achieved $96.7\%$ and $96.9\%$ accuracy with respect to the harvested apples. The main difference is the exclusion of ground apples in the yield data sets. To confirm the hypothesis that the ground apple patches are causing the drop in performance on test set 1, the yield data set should be manually labeled and used for evaluation.

To further improve counting results, our approach would benefit from a generic method to generate more training data synthetically. This would allow removing the up-sampled training images, containing three or more apples, from the training dataset. Another potential cause of errors are fruits on the ground. Those are not relevant to the overall yield estimate and should be discarded during the detection step. 
Our method should be tested on other types of fruits that grow in clusters, such as strawberries.

\bibliographystyle{IEEEtran}
\bibliography{IROS18}

\end{document}